\documentclass[10pt,twocolumn,letterpaper]{article}

\usepackage{cvpr}
\usepackage{times}
\usepackage{epsfig}
\usepackage{graphicx}
\usepackage{amsmath}
\usepackage{amssymb}
\usepackage{xcolor}
\usepackage{pifont}
\usepackage{overpic}
\usepackage{amssymb}
\usepackage{tabu}

\usepackage[pagebackref=true,breaklinks=true,letterpaper=true,colorlinks,bookmarks=false]{hyperref}

\newcommand{\point}[0]{\mathbf{p}}
\newcommand{\latent}[0]{\mathbf{z}}
\newcommand{\fgrid}[0]{\mathbf{F}}

\newcommand{\cmark}{\ding{51}}%
\newcommand{\xmark}{\ding{55}}%

\makeatletter
\newcommand*{\defeq}{\mathrel{\rlap{%
                     \raisebox{0.3ex}{$\m@th\cdot$}}%
                     \raisebox{-0.3ex}{$\m@th\cdot$}}%
                     =}
\makeatother

\cvprfinalcopy 

\def\cvprPaperID{6068} 

\ifcvprfinal\fi
\begin{document}

\title{Implicit Functions in Feature Space for 3D Shape\\ Reconstruction and Completion}

\author{Julian Chibane\textsuperscript{1,2}
\and
Thiemo Alldieck\textsuperscript{1,3}
\and
Gerard Pons-Moll\textsuperscript{1}}

\makeatletter
\let\@oldmaketitle\@maketitle
\renewcommand{\@maketitle}{
	\@oldmaketitle
	\centering
	\vspace{-5mm}
	{\small \textsuperscript{1}Max Planck Institute for Informatics, Saarland Informatics Campus, Germany}\\
    {\small	\textsuperscript{2}University of W{\"u}rzburg, Germany}\\
    {\small	\textsuperscript{3}Computer Graphics Lab, TU Braunschweig, Germany}\\
    {\tt\scriptsize \{jchibane,gpons\}@mpi-inf.mpg.de alldieck@cg.cs.tu-bs.de} \\
    \vspace*{3.5mm}
	\includegraphics[width=0.95\textwidth]{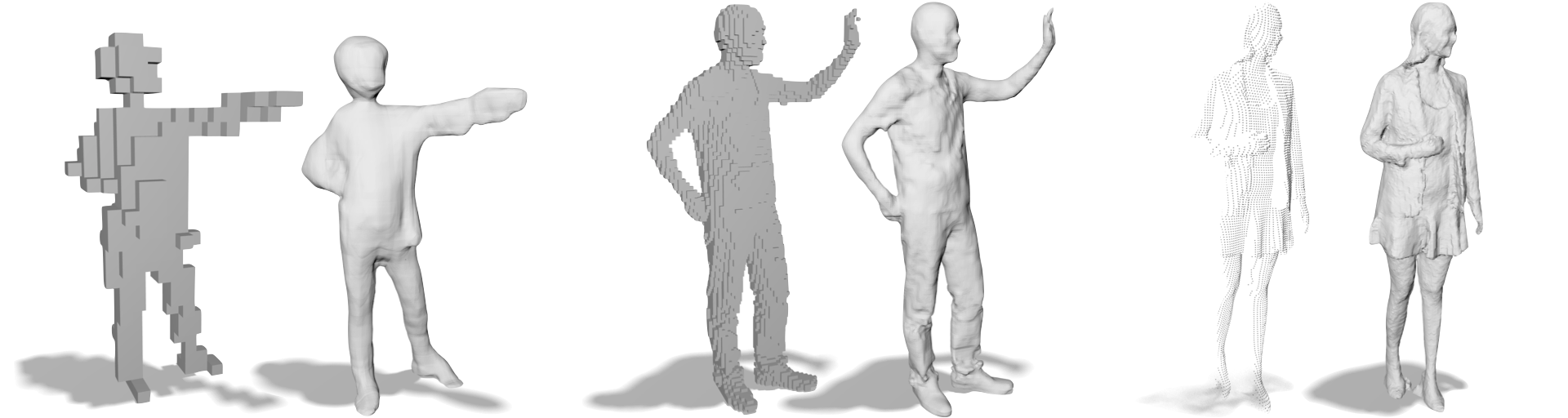}\\
	\refstepcounter{figure}\normalfont\small Figure~\thefigure: Results using our method. Left: sparse voxel reconstruction, middle: dense voxel reconstruction, right: 3D single-view point cloud reconstruction (back occluded). 
Our method delivers continuous outputs, handles multiple topologies (right) and unlike prior work, retains detail in the input (middle and right), and performs well with articulated humans.
	\label{fig:teaser}
	\vspace{5mm}
}
\makeatother

\maketitle
\thispagestyle{empty}



\begin{abstract}
\vspace{-4mm}
While many works focus on 3D reconstruction from images, in this paper, we focus on 3D shape reconstruction and completion from a variety of 3D inputs, which are deficient in some respect: 
low and high resolution voxels, sparse and dense point clouds, complete or incomplete. Processing of such 3D inputs is an increasingly important problem as they are the output of 3D scanners, which are becoming more accessible, and are the intermediate output of 3D computer vision algorithms. 
Recently, learned implicit functions have shown great promise as they produce continuous reconstructions.
However, we identified two limitations in reconstruction from 3D inputs: 
1) details present in the input data are not retained, and 2) poor reconstruction of articulated humans.
To solve this, we propose \textit{Implicit Feature Networks} (IF-Nets), which deliver continuous outputs, can handle multiple topologies, and complete shapes for missing or sparse input data retaining the nice properties of recent learned implicit functions, 
but critically they can also retain detail when it is present in the input data, and can reconstruct articulated humans.
Our work differs from prior work in two crucial aspects.
First, instead of using a single vector to encode a 3D shape, we extract a learnable 3-dimensional multi-scale tensor of deep features, which is aligned with the original Euclidean space embedding the shape. Second, instead of classifying x-y-z point coordinates directly, we classify deep features extracted from the tensor at a continuous query point.
We show that this forces our model to make decisions based on global and local shape structure, as opposed to point coordinates, which are arbitrary under Euclidean transformations. 
Experiments demonstrate that IF-Nets clearly outperform prior work in 3D object reconstruction in ShapeNet, and obtain significantly more accurate 3D human reconstructions. Code is available at \href{https://virtualhumans.mpi-inf.mpg.de/ifnets/}{https://virtualhumans.mpi-inf.mpg.de/ifnets/}.
\vspace{-6mm}
\end{abstract}

\section{Introduction}
\label{sec:introduction}
While many works focus on image-based 3D reconstruction~\cite{3Dobject_survey}, in this paper, we focus on 3D surface reconstruction and shape completion from a variety of 3D inputs, which are deficient in some respect: low-resolution voxel-grids, high-resolution voxel-grids, sparse and dense point-clouds, complete or incomplete. Such inputs are becoming ubiquitous as 3D scanning technology is increasingly accessible, and they are often an intermediate output of 3D computer vision algorithms. 
However, the final output for most applications should be a renderable continuous and complete surface, which is the focus of our work.




For sparse grids and (incomplete) point clouds, learning-based methods are a better choice than classical methods~\cite{c_ball_pivoting,c_marching_cubes}, as they reason about global object shape, but are limited by their output representation.
Mesh-based methods typically learn to deform an initial convex template~\cite{m_Wang18}, and hence can not represent different topologies.
Voxel-based representations~\cite{v_Choy16,v_Liao} have a large memory footprint, which critically limits the output resolution to coarse shapes, without detail. 
Point cloud~\cite{p_PointNet,p_PointNet++} representations are more efficient but do not trivially enable rendering and visualization of the surfaces.

Recently, implicit functions~\cite{i_DeepSDF,i_OccNet19,i_IMGAN19} have shown to be a promising shape representation for learning.
The key idea is to learn a function which, given a coarse shape encoded as a vector, and the x-y-z coordinates of a query point, decide whether the point is inside or outside of the shape. 
The learned implicit function can be evaluated at query 3D points at arbitrary resolutions, and the mesh/surface can be extracted applying the classical marching cubes algorithm. 
This output representation enables shape recovery at arbitrary resolutions, is continuous and can handle different topologies. 

While these approaches work well to reconstruct aligned rigid objects, we observed they suffer from two main limitations: 1) they can not represent complex objects like articulated humans (reconstructions often miss arms or legs), 2) they do not retain detail present in the input data. 
We hypothesize this occurs because 1) networks learn an overly strong prior on x-y-z point coordinates damaging the in-variance to articulation, and 2) the shape encoding vector lacks 3D structure, resulting in decodings that look more like classification into shape prototypes~\cite{v_ShapeReconstructionIsClassification} rather than continuous regression.
Consequently, all existing learning-based approaches, either based on voxels, meshes, points or implicit functions are lacking in some respect. 

In this paper we propose \emph{Implicit Feature Networks} (IF-Nets), which, unlike previous work, do well in 5 different axis, shown in Table~\ref{table:allmethods}: they are \emph{continuous}, can handle \emph{multiple topologies}, can complete data for \emph{sparse input}, retaining the nice properties of implicit function models~\cite{i_DeepSDF,i_OccNet19,i_IMGAN19}, but crucially they \emph{also} retain detail when is present in the input (\emph{dense input}), and can reconstructed \emph{articulated} humans. IF-Nets \emph{differ} from recent work~\cite{i_DeepSDF,i_IMGAN19,i_OccNet19} in two crucial aspects. First, instead of using a single vector to encode a 3D shape, we extract a 3-dimensional multi-scale tensor of deep features, which is aligned with original Euclidean space embedding the shape. Second, instead of classifying x-y-z point coordinates directly, we classify deep features extracted at continuous query points. 
Hence, unlike previous work, IF-nets do not \emph{memorize} common x-y-z locations, which are arbitrary under Euclidean transformations. Instead, they make decisions based on multi-scale features encoding local and global object shape structures around the point. 

To demonstrate the advantages of IF-Nets, first, we show that IF-Nets can reconstruct simple rigid 3D objects at better accuracy than previous methods. In ShapeNet~\cite{shapenet}, IF-Nets outperform the state-of-the-art results. 
For articulated humans, we train IF-Nets and related methods on a dataset of $1600$ humans in varied poses, shapes, and clothing. In stark contrast to recent work~\cite{i_DeepSDF,i_IMGAN19,i_OccNet19}, IF-Nets can reconstruct articulated objects globally without missing limbs, while recovering detailed structures such as cloth wrinkles. Quantitative and qualitative experiments validate that IF-Nets are more robust to articulations and produce globally consistent shapes without losing fine-scale detail. To encourage further research in 3D processing, learning, and reconstruction, we make IF-Nets publicly available at https://virtualhumans.mpi-inf.mpg.de/ifnets/.

\section{Related Work}
\label{sec:related}
\begin{table}
	\begin{center}
	\footnotesize
	\begin{tabular}{| l c c c c c  |}
		\hline
		\multicolumn{1}{|c}{Output}  & Continuous & Multiple & Sparse & Dense & Arti- \\
		\multicolumn{1}{|c}{3D Repr.} & Output & topologies & Input & Input & culated \\
		\hline
		Voxels & \xmark & \cmark & \cmark & \xmark & \cmark  \\
		Points & \xmark & \cmark & \cmark & \xmark & \cmark \\
		Meshes & \xmark & \xmark & \cmark & \xmark & \cmark  \\
		Implicit$^*$ & \cmark & \cmark & \cmark & \xmark & \xmark \\
		\hline
		Ours & \cmark & \cmark & \cmark & \cmark & \cmark  \\
		\hline
	\end{tabular}
    \end{center}
	\caption{Overview of strengths and weaknesses of recent 3D reconstruction approaches classified by their output representation.
	Voxels, point clouds, and meshes are non-continuous and suffer from discretization.
	Meshes additionally have fixed topologies, which limits the space of representable 3D shapes.
	Recent learned implicit functions$^*$~\cite{i_OccNet19,i_IMGAN19, i_DeepSDF} alleviate these limitations but fail to retain details or reconstruct articulation.
	The proposed IF-Nets share the desired properties of implicit functions for reconstructing from 3D inputs, but are \textit{additionally} able to preserve detail present in dense 3D input and to reconstruct articulated humans.}
	\label{table:allmethods}
\end{table}
Approaches for \emph{3D shape reconstruction} can be classified according to the representation used: voxels, meshes, point clouds, and implicit functions; and according to the object type: rigid objects vs humans.
For a more exhaustive recent review, we refer the reader to~\cite{3Dobject_survey}. A condensed overview of strengths and weaknesses of recent 3D reconstruction approaches is given in Table \ref{table:allmethods}.

\textbf{Voxels for rigid objects:} 
Since voxels are a natural 3D extension to pixels in image grids and admit 3D convolutions, they are most commonly used for generation and reconstruction~\cite{v_Kar17,v_Mengqi17,v_Jimenez16, v_olszewski2019tbn}. 
However, the memory footprint scales cubically with the resolution, which limited early works~\cite{v_Wu15,v_Choy16,v_Tulsiani17} to predict shapes in small $32^3$ grids.  Higher resolutions have been used~\cite{v_Wu16,v_Wu17,v_Zhang18} at the cost of limited training batches and slow training or lossy 2D projections~\cite{v_Smith18}. Multi-resolution~\cite{v_Hane17,v_Tatarchenko17,v_local_global} reconstruction reduced the memory footprint, allowing grids of size $256^3$. However, the approaches are complicated to implement, require multiple passes over the input, and are still limited to grids of size $256^3$, which result in visible quantization artifacts. To smooth out noise, it is possible to represent shapes as Truncated Signed Distance functions~\cite{v_TSDF_CurlessL96} for learning~\cite{v_Dai17,v_Ladicky17,v_Riegler17,v_stutz}. The resolution is however still bounded by the 3D grid storing the TSDF values.

Generative shape models typically map a $1D$ vector to a voxel representation with a neural network~\cite{v_Girdhar16,v_Wu16}. Like us, the authors of~\cite{v_Liu18} observe that the $1D$ vector is too restrictive to generate shapes with global and local structures. They introduce a hierarchical latent code with skip connections. Instead, we propose a much simpler 3-dimensional multi-scale feature tensor, which is aligned with original Euclidean space embedding the shape. 


\textbf{Humans with voxels:}
From images, CNN based reconstruction of humans represented as voxels~\cite{vh_varol2018bodynet,vh_Gilbert,vh_DeepHumans} or depth-maps~\cite{vh_Rogez,vh_facsimile,vh_Leroy18} typically produce more details than mesh or template-based representations, because predictions are aligned with the input pixels. Unfortunately, this comes at the cost of missing parts in the body. Hence, some methods~\cite{vh_varol2018bodynet,vh_facsimile} fit the SMPL~\cite{mh_smpl2015loper} model to the reconstructions as a post-processing step. This is however prone to fail if the original reconstructions are too incomplete. 
All these approaches process image pixels whereas we focus on processing 3D data directly. Unlike our IF-Nets, these methods are bounded by the resolution of the voxel grid. 

\textbf{Meshes for rigid objects:}
Most mesh-based methods predict shape as a deformation from a template~\cite{m_Wang18,m_Ranjan18} and hence are limited to a single topology. 
Alternatively, the mesh (vertices and faces) can be inferred directly~\cite{m_Groueix18,m_Dai19} -- while this research direction is promising, methods are still computationally expensive and can not guarantee a closed mesh without intersections. Direct mesh prediction can also be obtained using a learnable version~\cite{v_Liu18} of the classical marching cubes algorithm~\cite{c_marching_cubes}, but the approach is limited to an underlying small voxel grid of $32^3$. Promising combinations of voxels and meshes have been proposed~\cite{meshRCNN}, but results are still coarse. 

\textbf{Meshes for humans:}
Since the introduction of the (mesh-based) SMPL human model~\cite{mh_smpl2015loper} there have been a growing number of papers leveraging it to reconstructing shape and pose from point clouds, depth data and images~\cite{mh_kanazawa2018endtoend,mh_Pavlakos18,mh_Kolotouros19,mh_omran2018neural,mh_tung2017self,mh_zanfir2018monocular}. Since SMPL does not model clothing and detail, recent methods predict deformations from SMPL~\cite{mh_alldieck19cvpr,mh_alldieck2018detailed,mh_alldieck2018video,mh_ponsmoll2017clothcap,mh_bhatnagar2019mgn} or a template~\cite{mh_habermann2019TOG,mh_habermann20deepcap}. Unfortunately, CNN based mesh predictions tend to be over-smooth. More detail can be obtained predicting normals and displacement maps on a UV-map/geometry image of the surface~\cite{mh_alldieck2019tex2shape,mh_lazova3dv2019,mh_Pumarola}. However, all these approaches require different templates~\cite{mh_bhatnagar2019mgn,mh_patel2020} for every new garment topology or do not produce high-quality reconstructions~\cite{mh_Pumarola}.

\textbf{Point clouds for rigid objects:}
Processing point clouds is an important problem as they are the output of many sensors (LiDAR, 3D scanners) and computer vision algorithms. Due to their low weight, they have been also popular in computer graphics for representing and manipulating shapes~\cite{p_Pauly}. PointNet based architectures~\cite{p_PointNet,p_PointNet++} were the first to process point clouds directly for classification and semantic segmentation. The idea is to apply a fully connected network to each point followed by a global pooling operation to achieve permutation invariance. 
Recent architectures apply kernel point convolutions~\cite{p_Kernel}, tree-based graph convolutions~\cite{p_treeGAN}, and normalizing flows~\cite{p_PointFlow}. Point clouds are also used as shape representation for reconstruction~\cite{p_Fan17,p_InsafutdinovD18} and generation~\cite{p_PointFlow}. Unlike voxels or meshes, point clouds need to be non-trivially post-processed using classical methods~\cite{c_ball_pivoting,c_poission,c_poission2,c_Calakli11} to obtain renderable surfaces. 

\textbf{Point clouds for humans:}
Very few works represent humans with point clouds~\cite{ph_aumentado2019geometric}, probably because they can not be rendered. Recent works have employed either PointNet architectures~\cite{p_Jiang} or architectures based on point bases~\cite{p_humans_Basis} to register a human mesh to the point cloud. 

\textbf{Implicit Functions for rigid objects:}
Recently, neural networks have been used to learn a continuous implicit function representing shape~\cite{i_OccNet19,i_DeepSDF,i_IMGAN19,i_DeepLevelSet, i_StructuredImplicitFunctions}. For this a neural network can be feed with a latent code and a query point ($x$-$y$-$z$) to predict the TSDF value~\cite{i_DeepSDF} or the binary occupancy of the point~\cite{i_OccNet19,i_IMGAN19}. A recent method~\cite{i_DISN} achieved state-of-the-art results for 3D reconstruction from images combining 3D query point features with local image features, by approximating the projection of the query point onto the 2D image with a view-point prediction. This trick of querying continuous points used in implicit function learning allows predicting in continuous space (potentially at any resolution), breaking the memory barrier of voxel-based methods. 
These works inspired our work, but we note that they can not reconstruct articulated humans from 3D data: \cite{i_DISN} can not take 3D inputs as point clouds or voxel grids and relies on an approximate 3D to 2D projection losing details; the reconstructions of \cite{i_OccNet19,i_IMGAN19} often miss limbs. We hypothesize that \cite{i_OccNet19,i_IMGAN19,i_DISN} memorize point coordinates instead of reasoning about shape, and that the vectorized latent 1D vector representation~\cite{i_OccNet19,i_IMGAN19} is not aligned with the input, and lacks 3D structure. We address this issues by querying deep features extracted at continuous locations of a 3D grid of multi-scale features aligned with the 3D input space. This modification is easy to implement and results in significant gains in reconstruction quality.

\textbf{Implicit Functions for humans:}
TSDFs~\cite{v_TSDF_CurlessL96} have been used to represent human shapes for depth-fusion and tracking~\cite{ih_Dynamic_Fusion,ih_KillingFusion}. Such implicit representation has been combined with the SMPL~\cite{mh_smpl2015loper} body model to significantly increase tracking robustness and accuracy~\cite{ih_DoubleFusion}.
From an input image, humans in clothing are predicted using an implicit network~\cite{ih_PiFu}. \cite{ih_PiFu} produces higher quality results compared to prior implicit function work \cite{i_IMGAN19}. The reconstruction is done by pointwise occupancy prediction based on the location of a 3D query point and 2D image features. For simple poses, the approach produces very compelling and detailed results but struggles for more complex poses. 
The approach~\cite{ih_PiFu} does not incorporate a multi-scale 3D shape representation like ours, and it is designed for image reconstruction, whereas we focus on 3D reconstruction from sparse and dense point clouds and occupancy grids. 
Like previous implicit networks, our method produces continuous surfaces at arbitrary resolution. 
But importantly, by virtue of our 3D multi-scale shape representation aligned with the input space, our reconstructions preserve global structure while retaining fine-scale detail, even for complex poses. 


\section{Method}
\label{sec:method}
\begin{figure*}
\begin{center}
\begin{overpic}[width=1\linewidth]{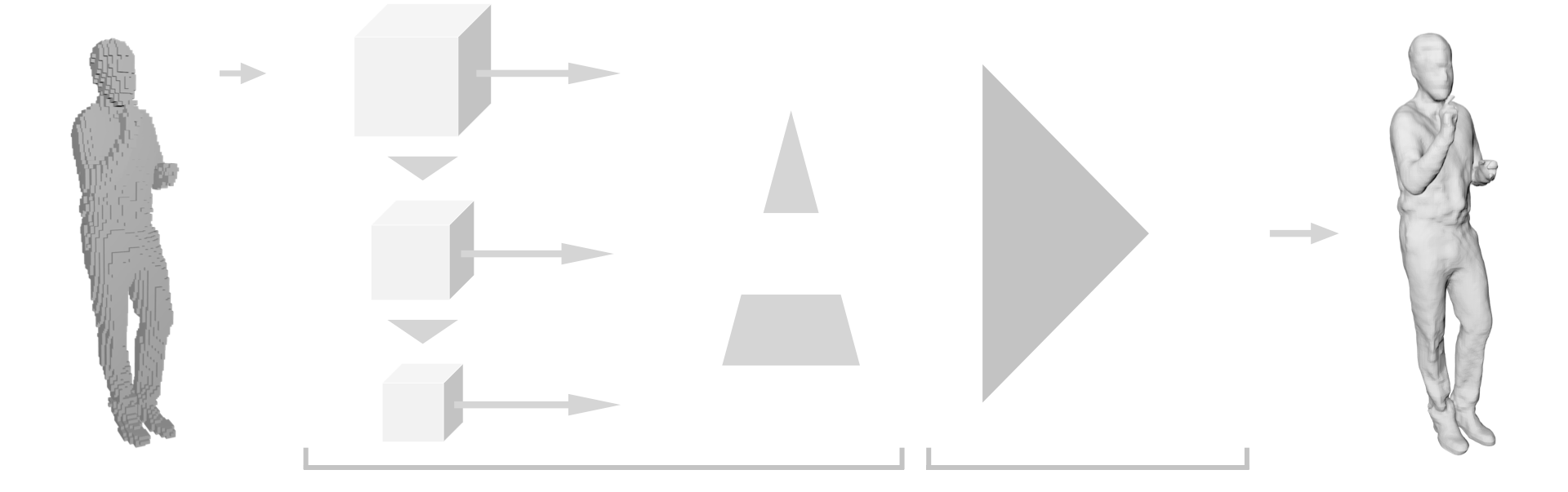} 
    \put(18,26){\large$\fgrid_1$}
    \put(18,14.2){\large$\fgrid_2$}
    \put(19,9.3){\rotatebox{90}{\dots}}
    \put(18,4.8){\large$\fgrid_n$}
    
    \put(24,26.5){$\bigotimes_\point$}
    \put(24.5,14.5){\small$\bigotimes_\point$}
    \put(25,4.8){\footnotesize$\bigotimes_\point$}
    
    \put(47.5,26){\large$\fgrid_1(\point)$}
    \put(47.5,14.2){\large$\fgrid_2(\point)$}
    \put(50.2,9.3){\rotatebox{90}{\dots}}
    \put(47.5,4.8){\large$\fgrid_n(\point)$}
    
    \put(56.5,20){\footnotesize \emph{local}}
    \put(56,10){\footnotesize \emph{global}}
    
    
    \put(65,15.5){\Huge $f$}
    \put(74.6,15.9){\large $[0 , 1]$}
    
    \put(7,-0.5){Input}
    \put(87.2,-0.5){Reconstruction}
    
    \put(27,-0.5){\footnotesize learned, multi-scale point encoding}
    \put(58.5,-0.5){\footnotesize learned point occupancy decoding}
    \put(13,16.2){\Huge $g$}

\end{overpic}
\end{center}
   \caption{Overview of IF-Nets: given an (incomplete or low resolution) input, we compute a 3D grid of multi-scale features, encoding global and local properties of the input shape. 
   	Then, we extract deep features $\fgrid_1(\mathbf{p}) \hdots \fgrid_n(\mathbf{p})$ from the grid at \emph{continuous point} locations $\point$. Based \emph{only on these features} a decoder $f(\cdot)$ decides whether the point $\point$ lies inside (classification as $1$) or outside (classification as $0$) the surface. Like recent implicit function-based works, we can query at arbitrary resolutions and reconstruct a continuous surface.
	Unlike them, our method reasons based exclusively on point-wise deep features, instead of point coordinates. This allows us to reconstruct articulated structures and preserve input detail. }
\label{fig:method_overwiew}
\vspace{-2mm}
\end{figure*}

To motivate the design of our Implicit Feature Networks (IF-Nets), we first describe the formulation of recent learned implicit functions, pointing out their strengths and weaknesses in Sec.~\ref{subsec:background}. We explain our IF-Nets in Sec.~\ref{subsec:ifnets}. The key ideas of IF-Nets are illustrated in Fig.~\ref{fig:method_overwiew}.

\subsection{Background: Learning with Implicit Surfaces}
\label{subsec:background}

While recent works~\cite{i_DeepSDF,i_OccNet19,i_IMGAN19} on learned implicit reconstruction from 3D input differ in their inference and their output shape representation (signed distance or binary occupancies), they are conceptually similar. Here, we describe the occupancy formulation of~\cite{i_OccNet19}. Note that the strengths and limitations of these methods are very similar. 
They all encode a 3D shape using a latent vector $\latent \in \mathcal{Z} \subset \mathbb{R}^m$. 
Then a continuous representation of the shape is obtained by learning a neural function
\begin{equation}
  f(\mathbf{z},\point) \colon \mathcal{Z} \times \mathbb{R}^3 \mapsto [ 0,1 ],
\label{eq:implicit}
\end{equation}
which given a query point $\point \in \mathbb{R}^3$, and the latent code $\latent$, classifies whether the point is inside (classification as 1) or outside (classification as 0) the surface.
Thereby, the surface is implicitly represented as the points on the decision boundary, $\{\point \in \mathbb{R}^3 \,| \,f(\latent,\point) = t\}$, with a threshold parameter $t$ ($t=0.5$ for IF-Nets).

Once $f(\cdot)$ is learned, it can be queried at continuous point locations, without resolution restrictions imposed by typical voxel grids.
To construct a mesh, marching cubes~\cite{c_marching_cubes} can be applied on the predicted occupancy grid. 
This elegant formulation breaks the barriers of previous representations allowing detailed reconstruction of complex topologies, and
 has proven effective for several tasks such as rigid object reconstruction from images, occupancy grids, and point clouds. 
However, we observed that models of this kind suffer from two main limitations: 1) they can not represent complex objects like articulated objects, and 2) they do not preserve detail present in the input data. 
We address these limitations with IF-Nets. 

\subsection{Implicit Feature Networks}
\label{subsec:ifnets}
We identify two potential problems with the previous formulation. First, directly inputting point coordinates $\point$ gives the network the option to by-pass reasoning about shape structure, by memorizing typical point occupancies for object prototypes. This severely damages reconstruction in-variance to rotation and translation, which is one of the cornerstones of successful 2D convolution networks for segmentation, recognition, and detection. Second, encoding the full shape in a single vector $\latent$ loses detail present in the data, and loses alignment with the original 3D space where shapes are embedded. 


In this work, we propose a novel encoding and decoding tandem capable of addressing the above limitations for the task of 3D reconstruction from point clouds or occupancy grids. 
Given such 3D input data $\mathbf{X} \in \mathcal{X}$ of an object, where $\mathcal{X}$ denotes the space of the inputs, and a 3D point $\point \in \mathbb{R}^3$, we want to predict if $\point$ lies inside or outside the object. 

\paragraph{Shape Encoding:} Instead of encoding the shape in a single vector $\mathbf{z}$, we construct a rich encoding of the data $\mathbf{X}$ through subsequently convolving it with learned 3D convolutions. This requires the input to lie on a discrete voxel grid, i.e. $\mathcal{X} = \mathbb{R}^{N \times N \times N}$, where $N \in \mathbb{N}$ denotes the input resolution. To process point clouds we simply discretize them first.  
The convolutions are followed by down scaling the input, creating growing receptive fields and channels but shrinking resolution, just like commonly done in 2D~\cite{imagenet}. Applying this procedure recursively $n$ times on the input data $\mathbf{X}$, we create \emph{multi-scale deep feature grids} $\fgrid_1,..,\fgrid_n$, $\fgrid_k \in \mathcal{F}_k^{K\times{K}\times{K}}$, of decreasing resolution $K = \frac{N}{2^{k-1}}$, and variable channel dimensionality $F_k \in \mathbb{N}$ at each stage $\mathcal{F}_k \subset \mathbb{R}^{F_k}$. 
The feature grids $\mathbf{F}_k$ at the early stages (starting at $k=1$) capture high frequencies (shape detail), whereas feature grids $\mathbf{F}_k$ at the late stages (ending at stage $k=n$) have a large receptive fields, which capture the global structure of the data. This enables to reason about missing or sparse data, while retaining detail when is present in the input. We denote the encoder as \begin{equation}
g(\mathbf{X}) \defeq \fgrid_1,..,\fgrid_n \ .
\label{eq:encoder}
\end{equation}

\paragraph{Shape Decoding:} Instead of classifying point coordinates $\point$ directly, we extract the learned deep features $\fgrid_1(\point),..,\fgrid_n(\point)$ from the feature grids at location $\point$. This is only possible because our \emph{encoding has a 3D structure} aligned with the input data. Since feature grids are discrete, we use trilinear interpolation to query continuous 3D points $\point \in \mathbb{R}^3$. 
In order to encode information of the local neighborhood into the point encoding, even at early grids with small receptive fields (e.g.\ $\mathbf{F}_1$), we extract features at the location of a query point $\mathbf{p}$ itself and \textit{additionally} at surrounding points in a distance $d$ along the Cartesian axes:
\begin{equation}
\{\point + a\cdot \mathbf{e}_i \cdot d \in \mathbb{R}^3 | a \in \{1,0,-1\}, i \in \{1,2,3\}\},
\end{equation}
where $d \in \mathbb{R}$ is the distance to to the center point $\point$ and $\mathbf{e}_i \in \mathbb{R}^3$ is the $i-th$ Cartesian axis unit vector, see supplementary material for an illustration.

The point encoding $\fgrid_1(\point),..,\fgrid_n(\point)$, with $\fgrid_k(\point) \in \mathcal{F}_k$, is then fed into a point-wise decoder $f(\cdot)$, parameterized by a fully connected neural network, to predict if the point $\point$ lies inside or outside the shape: 
\begin{equation}
f(\fgrid_1(\point),\hdots, \fgrid_n(\point))  \colon \mathcal{F}_1 \times \hdots \times \mathcal{F}_n \mapsto [0,1]
\label{eq:ifnet}
\end{equation}
In contrast to Eq.~\eqref{eq:implicit}, in this formulation, the network classifies the point based on local and global shape features, instead of point coordinates, which are arbitrary under rotation, translation, and articulation transformations. Furthermore, due to our multi-scale encoding, details can be preserved while reasoning about global shape is still possible.

\subsection{Method Training}
\label{sec:method_training}
To train the multi-scale encoder $g_{\mathbf{w}}(\cdot)$ in Eq.~\eqref{eq:encoder}, and decoder $f_{\mathbf{w}}(\cdot)$ in Eq.~\eqref{eq:ifnet}, parameterized with neural weights $\mathbf{w}$, pairs  $\{\mathbf{X}_i, \mathcal{S}_i\}_{i=1}^T$ of 3D inputs $\mathbf{X}_i$ with corresponding 3D ground truth object surfaces $\mathcal{S}_i$ are required, where $i \in 1,\dots, T$ and $T \in \mathbb{N}$ denotes the number of such training examples. The notation $g_{\mathbf{w}}(\mathbf{X},\point) \defeq \fgrid_1^\mathbf{w}(\point),\hdots, \fgrid_n^\mathbf{w}(\point)$ denotes evaluation of the multi-scale encoding at point $\point$. 
To create training point samples, for every ground truth surface $\mathcal{S}_i$, we sample a number $S \in \mathbb{N}$ of points $\point_i^j \in \mathbb{R}^3$, $j \in 1, \dots, S$. To this end, we first make the ground truth surface $\mathcal{S}_i$ watertight. Then we compute the ground truth occupancy $o_i(\point_i^j) \in \{0,1\}$, which evaluates to $1$ for inside points and $0$ otherwise. Next, the point samples $\point_i^j$ are created near the surface by sampling points $\point_{i,j}^\mathcal{S} \in \mathcal{S}_i$ on the ground truth surfaces and adding random displacements $\mathbf{n}_{i,j} \sim \mathcal{N}(0, \boldsymbol{\Sigma})$, i.e. $\point_i^j \defeq \point_{i,j}^\mathcal{S} + \mathbf{n}_{i,j}$. To this end, we use a diagonal covariance matrix $\boldsymbol{\Sigma} \in \mathbb{R}^{3\times{3}}$ with entries $\boldsymbol{\Sigma}_{i,i} = \sigma$. We find good results by sampling $50\%$ of the point samples very near the surface with a small $\sigma_1$, and $50\%$ in the further away surroundings with a larger $\sigma_2$. For training, the network weights $\mathbf{w}$ are optimized by minimizing the mini-batch loss
\begin{align}
\small
\label{eq:loss}
\mathcal{L}_\mathcal{B}({\mathbf{w}}) 
&\defeq \sum_{i \in \mathcal{B}} \sum_{j \in \mathcal{R}} L (f_{\mathbf{w}}(g_{\mathbf{w}}(\mathbf{X}_i, \point_i^j)), o_i(\point_i^j)) \\
&= \sum_{i \in \mathcal{B}} \sum_{j \in \mathcal{R}} L (f_{\mathbf{w}}(\fgrid_1^\mathbf{w}(\point_i^j),\hdots, \fgrid_n^\mathbf{w}(\point_i^j)), o_i(\point_i^j)) \nonumber,
 \label{eq:loss}
\end{align} 
which sums over training surfaces $i \in \mathcal{B} \subset 1, \dots, T$ of a given mini-batch $\mathcal{B}$ and point samples $j \in \mathcal{R} \subset 1, \dots, S$ of a subsample $\mathcal{R}$. The subsample $\mathcal{R}$ is regenerated for every evaluation of the mini-batch loss $\mathcal{L}_\mathcal{B}$. For $L(\cdot,\cdot)$, we use the standard cross-entropy loss.
By minimizing $\mathcal{L}_\mathcal{B}$, we train the encoder $g_{\mathbf{w}}(\cdot)$ and the decoder $f_{\mathbf{w}}(\cdot)$ jointly and end-to-end.
Please see the supplementary material for the concrete values for the hyperparameters used in the experiments.

\subsection{Method Inference}
At test time, the goal is to reconstruct a continuous and complete representation, given only a discrete and incomplete 3D input $\mathbf{X}$. First, we use the learned encoder network to construct the multi-scale feature grids $g(\mathbf{X}) = \fgrid_1,..,\fgrid_n$.
Then, we use the point-wise decoder network $f(g(\mathbf{X},\point))$ to create occupancy predictions at continuous point locations $\point \in \mathbb{R}^3$ (cf.\ Sec.~\ref{subsec:ifnets}). In order to construct a mesh, we evaluate the IF-Net on points on a grid of the desired resolution. Then, the resulting high resolution occupancy grid is transformed into a mesh using the classical marching cubes~\cite{c_marching_cubes} algorithm.

 \section{Experiments}
\label{sec:experiments}
In this section we validate the effectiveness of IF-Nets on the challenging task of 3D shape reconstruction. We show that our IF-Nets are able to address two limitations of recent learning-based approaches for this task: 1) IF-Nets preserve detail present in the input data, while also reasoning about incomplete data, 2) IF-Nets are able to reconstruct articulated humans in complex clothing. To this end, we conduct three experiments of increasing complexity: \emph{Point Cloud Completion} (Sec.~\ref{subsec:pointcloud_completion}), \emph{Voxel Super-Resolution} (Sec.~\ref{subsec:voxel_super_resolution}) and \emph{Single-View Human Reconstruction} (Sec.~\ref{subsec:svhr}).



\textbf{Baselines:}
For the task of Point Cloud Completion, we evaluate our approach against Occupancy Networks~\cite{i_OccNet19} (OccNet), Point Set Generation Networks~\cite{p_Fan17} (PSGN) and Deep Marching Cubes~\cite{v_Liao} (DMC). For Voxel Super-Resolution, we compare against IMNET \cite{i_IMGAN19}  as well as again against OccNet and DMC. 
For DMC and PSGN we used the implementations provided online by the authors of \cite{i_OccNet19}. We trained all methods until the validation minimum was reached. Training was repeated for every considered experiment setup. To show a consistent comparison, we modified the IMNET implementation to be able to be trained on all ShapeNet classes jointly. For IMNET and OccNet, we kept the sampling strategies proposed by their authors. For IMNET, we followed the authors and performed progressive resolution increasing of training data sampling during training. 

\textbf{Metrics:}
To measure reconstruction quality quantitatively, we consider three established metrics (see suppl. of \cite{i_OccNet19} for definition and implementation): volumetric \emph{intersection over union} (IoU) measuring how well the defined volumes match (higher is better), \emph{Chamfer-$L_2$} measuring the accuracy and completeness of the surface (lower is better), and \emph{normal consistency} measuring the accuracy and completeness of the shape normals (higher is better).

\textbf{Data:}
We consider two datasets: 1) a dataset containing 3D scans of humans\footnote{The dataset will be available for purchase from Twindom.} to assess the challenging task of reconstruction from incomplete and articulated shapes and 2) the established ShapeNet~\cite{shapenet} dataset, consisting of rigid object classes, with rather prototypical shapes like cars, airplanes, and rifles. The ShapeNet data has been pre-processed to be watertight by the authors of \cite{i_DISN}, allowing to compute ground truth occupancies and scaled such that every shape's largest bounding box edge has length one. We conduct all experiments and evaluations using pre-possessed ShapeNet data and use the common training and test split by \cite{v_Choy16}. However, preprocessing failed for some objects, leading to broken objects with large holes. Therefore, 508 heavily distorted objects have been removed for meaningful evaluation. The filtered list of all used objects is published alongside the code.
We also evaluate on a challenging dataset consisting of scanned humans in highly varying articulations with complex and varying clothing topologies like coats, skirts, or hats.
The scans have been captured using commercial 3D scanners.
The dataset, referred to as \emph{Humans}, consists of 2183 such scans, split into 478 examples for testing, 1598 for training and 197 for validation. The scans have been height normalized and centered, but in contrast to the ShapeNet objects, exhibit varying rotations.

\begin{figure}[t]
\centering
\footnotesize
\begin{tabu} to \linewidth {X[c]X[c]X[c]X[c]X[c]X[c]}
   Input & OccNet & PSGN & DMC & Ours & GT\\
\end{tabu}
\includegraphics[width=1\linewidth]{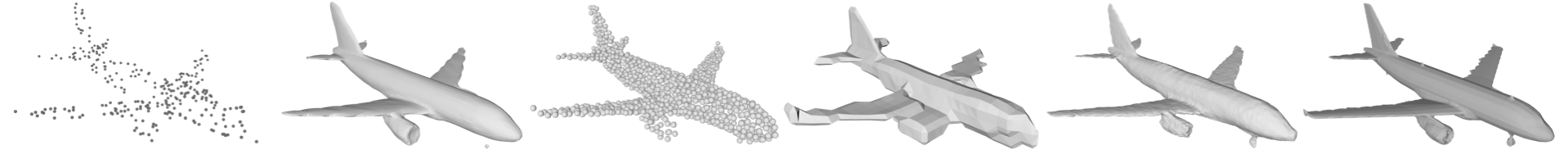}
\vspace{-3mm}
\\
\includegraphics[width=1\linewidth]{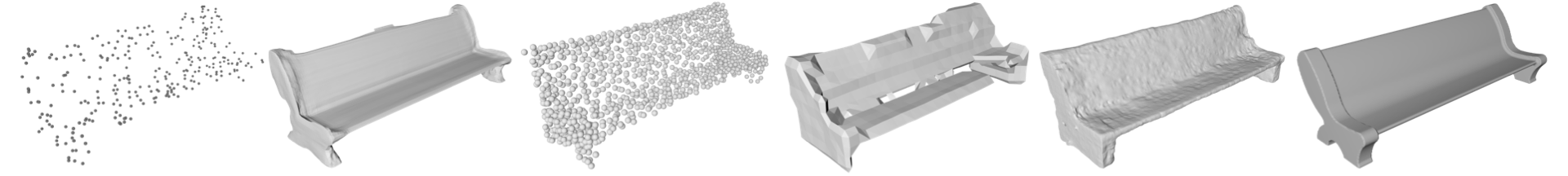}
\vspace{-3mm}
\\
\includegraphics[width=1\linewidth]{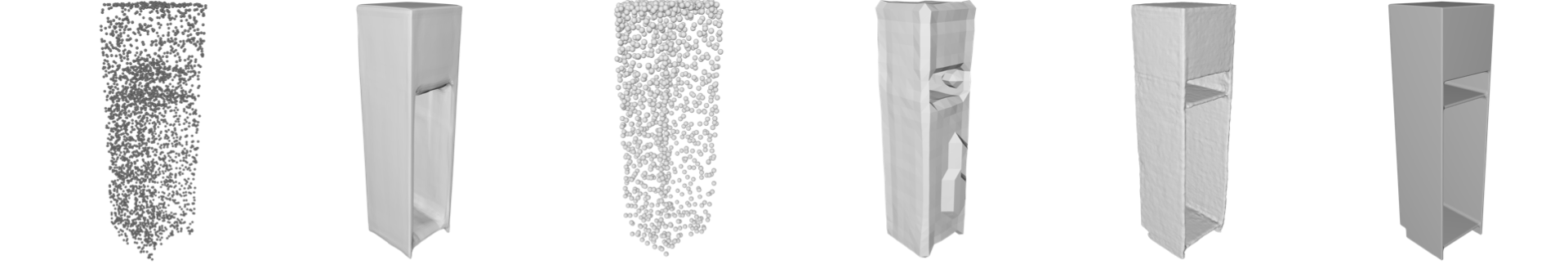}
\includegraphics[width=1\linewidth]{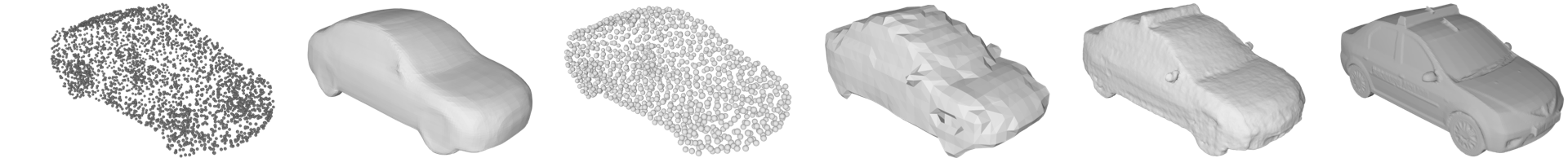}
\\\vspace{1mm}
\begin{tabu} to \linewidth {X[c]X[c]X[c]X[c]X[c]X[c]}
   Input & OccNet & IMNET & DMC & Ours & GT\\
\end{tabu}
\includegraphics[width=1\linewidth]{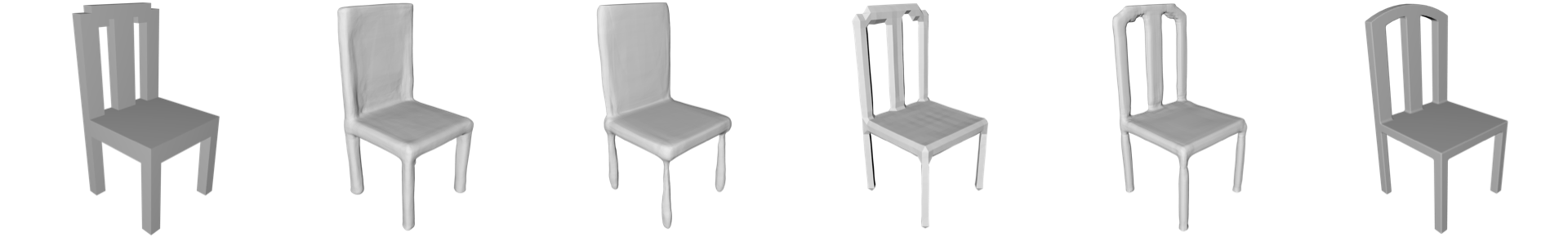}
\includegraphics[width=1\linewidth]{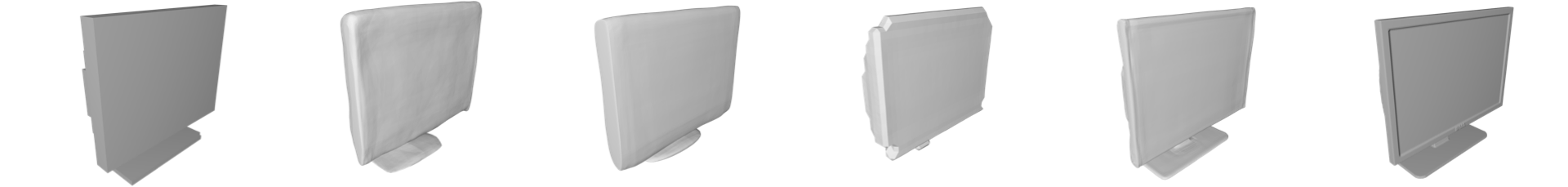}
\includegraphics[width=1\linewidth]{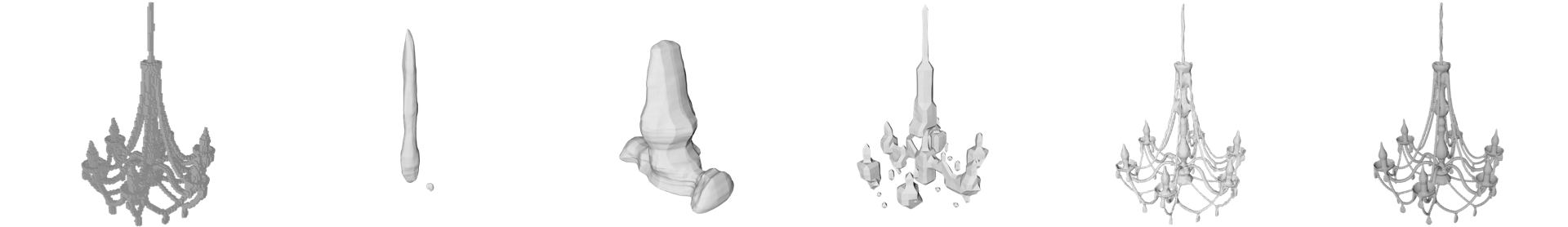}
\includegraphics[width=1\linewidth]{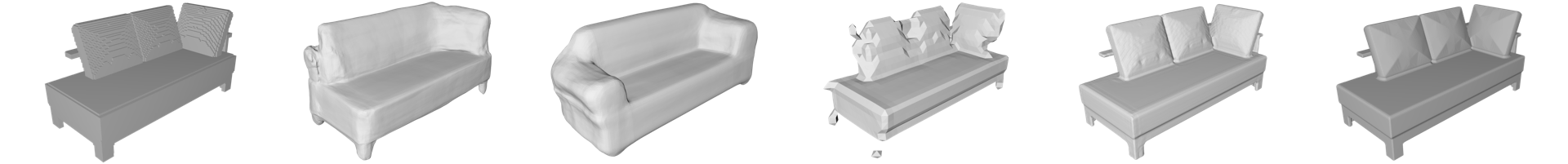}
\vspace{-2mm}
\caption{Qualitative results for two input types: point cloud (top) and voxels (bottom) on ShapeNet dataset. Each type is further subdivided into sparse (top two rows) and dense (bottom two rows).  }
\label{fig:shapenet}
\vspace{-3.5mm}
\end{figure}

\subsection{Point Cloud Completion}
\label{subsec:pointcloud_completion}
\begin{table}
	\footnotesize
\begin{center}




\begin{tabular}{l||c|c||c|c||c|c}

 & \multicolumn{2}{c||}{IoU $\uparrow$}   & \multicolumn{2}{c||}{Chamfer-$L_2$ $\downarrow$}  & \multicolumn{2}{c}{Normal-Consis. $\uparrow$} \\ \hline \hline 
Input  &  $-$  &  $-$ &    0.07   &    0.009           &  $-$  &  $-$    \\ 
OccNet &  0.73     &   0.72   &    0.03   &      0.04         &  0.88  &   0.88   \\ 

DMC   &  0.58      &  0.65   &     0.03       &     0.01          &     0.83   &   0.86   \\ 

PSGN   &  $-$  &  $-$ &   0.04  &  0.04     &     $-$     &     $-$     \\ 

Ours &    \textbf{0.79}    &   \textbf{0.88}    &     \textbf{0.02}   &   \textbf{0.002}    &      \textbf{0.90}      &    \textbf{0.95}    \\ 

\end{tabular}
\end{center}
\vspace{-1mm}
\caption{Results of point cloud reconstruction on ShapeNet. Left number indicates score from 300 points, right one from 3000 points. Chamfer-$L_2$ results $\times 10^{-2}$.}
\label{tab:shapenet_numbers_points}
\vspace{-1mm}
\end{table}
As a first task, we apply IF-Nets to the problem of completing sparse and dense point clouds -- we sample 300 points (sparse) and 3000 points (dense) respectively from ShapeNet surface models and ask our method to complete the full surfaces.
Completing point clouds is challenging since it requires to simultaneously preserve input details and reason about missing structure at the same time.
In Fig.~\ref{fig:shapenet} we show comparisons against the baseline methods.
Our method outperforms all baselines both in preserving local detail and recovering global structures.
For the dense point clouds, the strengths of our method are paramount.
Our method is the only one capable of reconstructing the car rear-view mirrors and the additional shelf of the wardrobe.
We additionally quantitatively compare our method and report the numbers in Tab.~\ref{tab:shapenet_numbers_points}.
Our method beats the state-of-the-art in all metrics by large margin. In fact, using 3000 points as input, all competitors produce results which have larger Chamfer distance than the input itself, suggesting they fail at preserving input detail.
Only IF-Nets preserve input details while completing missing structures.

\subsection{Voxel Super-Resolution}
\label{subsec:voxel_super_resolution}
\begin{table}
	\footnotesize
\begin{center}







\begin{tabular}{l||c|c||c|c||c|c}

 & \multicolumn{2}{c||}{IoU $\uparrow$}   & \multicolumn{2}{c||}{Chamfer-$L_2$ $\downarrow$}  & \multicolumn{2}{c}{Normal-Consis. $\uparrow$} \\ \hline \hline 
Input &    0.49   &  0.79    &     0.04       &    0.003     &  0.81      &   0.87        \\  
DMC   &    0.59   &   0.67    &    0.45      &     0.45     &    0.83    &    0.84        \\ 
IMNET   &   0.49   &   0.40   &    0.47       &     0.40     &     0.79    &     0.77       \\            
OccNet   &   0.60     &   0.71     &    0.10        &     0.05     &     0.85        &      0.88     \\ 
Ours &    \textbf{0.73}      &    \textbf{0.92}           &   \textbf{0.02}      &      \textbf{0.002}          &   \textbf{ 0.91 }       &   \textbf{0.98}    \\ 
\end{tabular}

\end{center}
\vspace{-1mm}
\caption{Results of voxel grid reconstruction on ShapeNet. For each metric, left column indicates score from $32^3$ resolution, right one from $128^3$ resolution. Chamfer-$L_2$ results $\times 10^{-2}$.}
\label{tab:shapenet_numbers_voxels}
\vspace{-1mm}
\end{table}
\begin{table}
	\footnotesize
\begin{center}






\begin{tabular}{l||c|c||c|c||c|c}

 & \multicolumn{2}{c||}{IoU $\uparrow$}   & \multicolumn{2}{c||}{Chamfer-$L_2$ $\downarrow$}  & \multicolumn{2}{c}{Normal-Consis. $\uparrow$} \\ \hline \hline 

Input   &  0.49    &   0.76     &    0.04         &    0.003        &      0.82        &     0.86 \\ 
 
DMC    &  0.77 & 0.85   &  0.03      & 0.01        &   0.79      &   0.83        \\

IMNET   & 0.63 & 0.64    &    0.27   & 0.23         &    0.79         & 0.79     \\

OccNet   &  0.63 & 0.65     &   0.22  & 0.19             &       0.79 & 0.79             \\

Ours &   \textbf{0.80}     & \textbf{0.96}  &   \textbf{0.02 } &  \textbf{0.001 }   &     \textbf{0.86}    & \textbf{ 0.94}    
\end{tabular}

\end{center}
\vspace{-1mm}
\caption{Results of voxel grid reconstruction on the Humans dataset. Left number indicates score from $32^3$ resolution, right one from $128^3$ resolution. Chamfer-$L_2$ results $\times 10^{-2}$. IF-Nets coherently outperform others in the incomplete data setup. IF-Nets show a large increase in performance with dense data, whereas others show similar performance. This demonstrates that IF-Nets are the first learned approach, to our knowledge, being able to faithfully reconstruct dense information present in 3D data.}
\vspace{-3mm}
\label{tab:human_numbers_voxels}
\end{table}

\begin{figure}[t]
\centering
\footnotesize
   \includegraphics[width=1\linewidth]{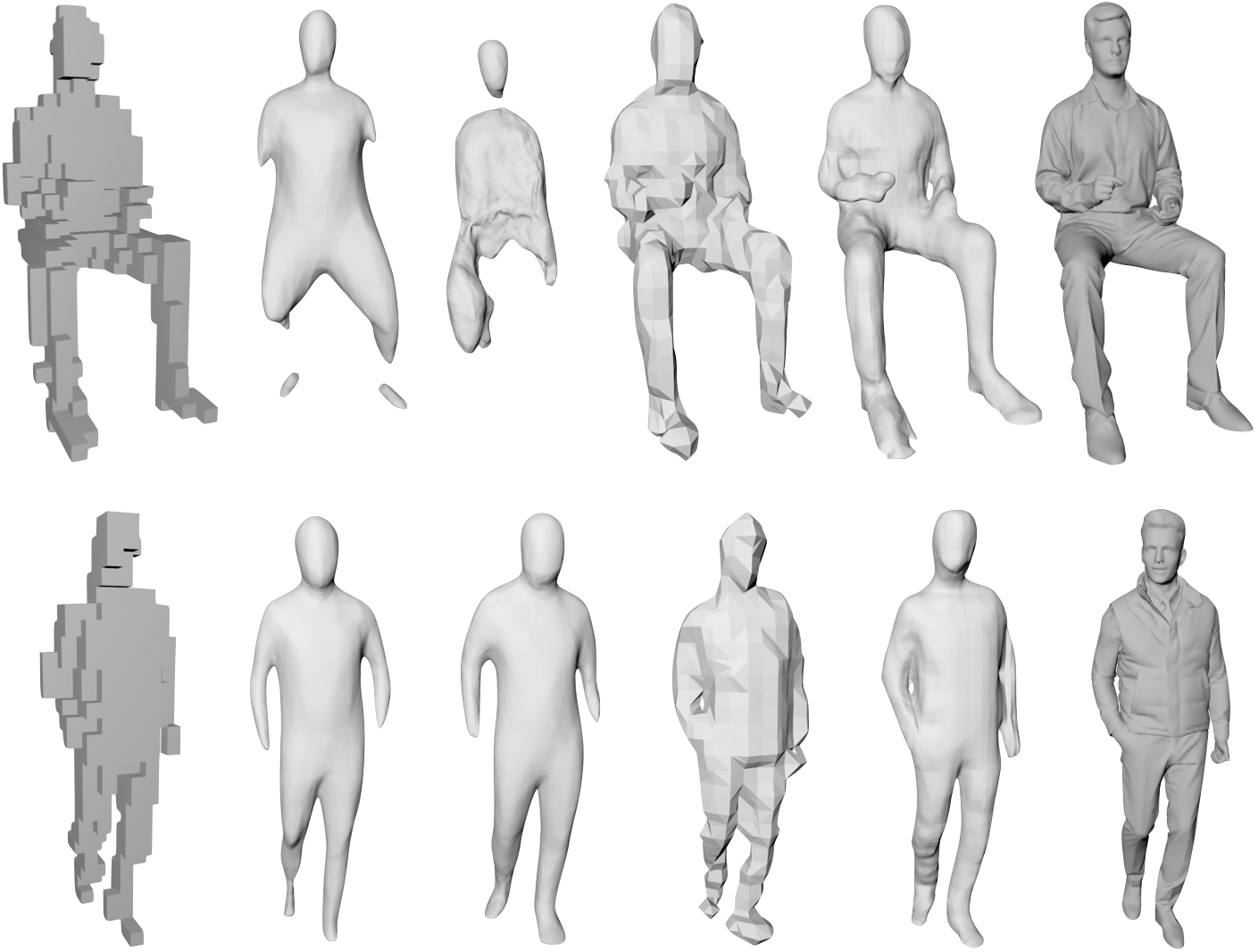}
   \includegraphics[width=1\linewidth]{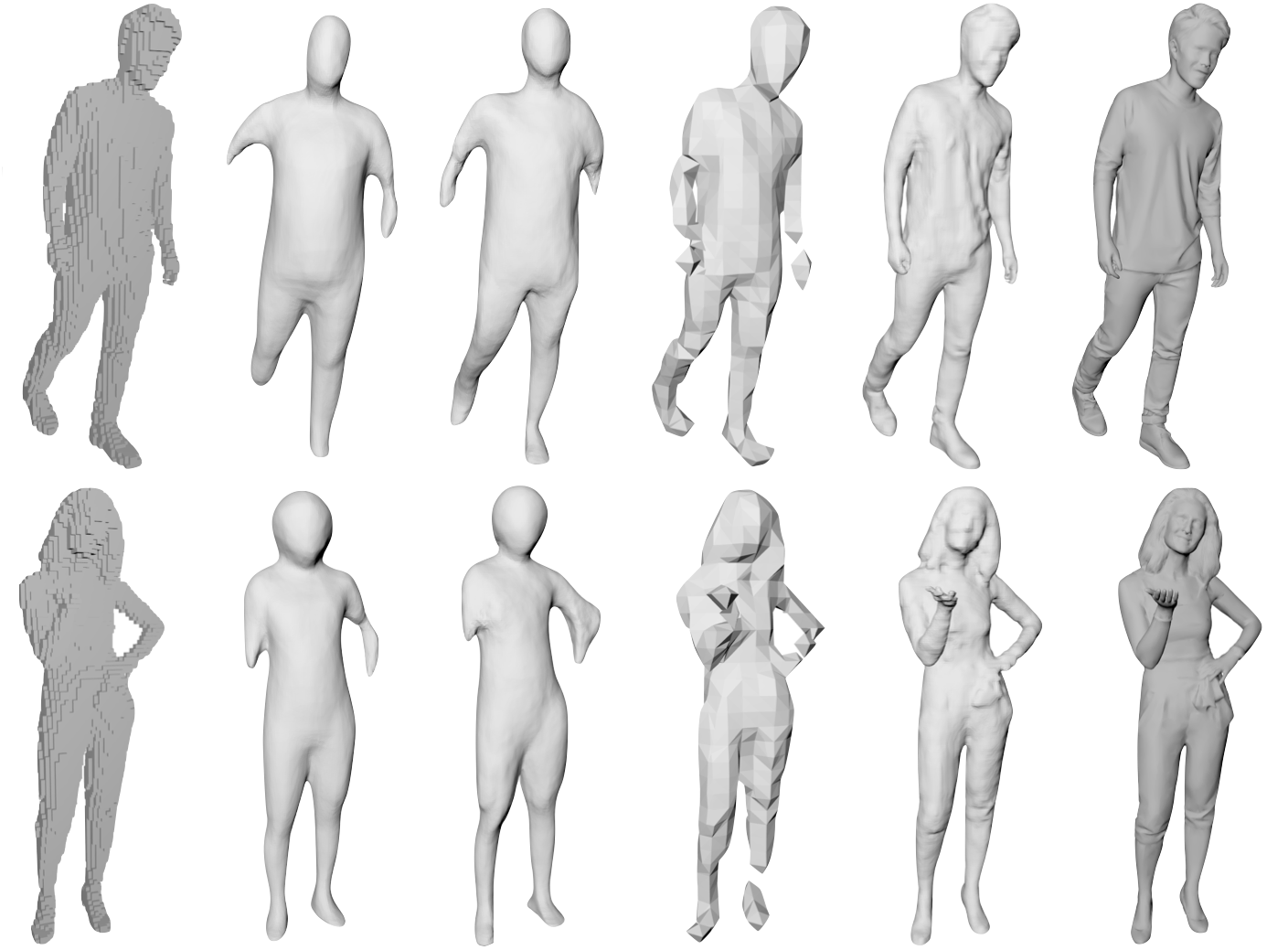}
\begin{tabu} to \linewidth {X[c]X[c]X[c]X[c]X[c]X[c]}
   Input & OccNet & IMNET & DMC & Ours & GT\\
\end{tabu}
\vspace{-2mm}
\caption{Qualitative results of sparse ($32^3$, upper) and dense ($128^3$, lower) 3D voxel super-resolution on the Humans dataset.}
\vspace{-3.5mm}
\label{fig:voxel_humans}
\end{figure}

\begin{figure*}[ht]
\centering
\includegraphics[width=0.95\linewidth]{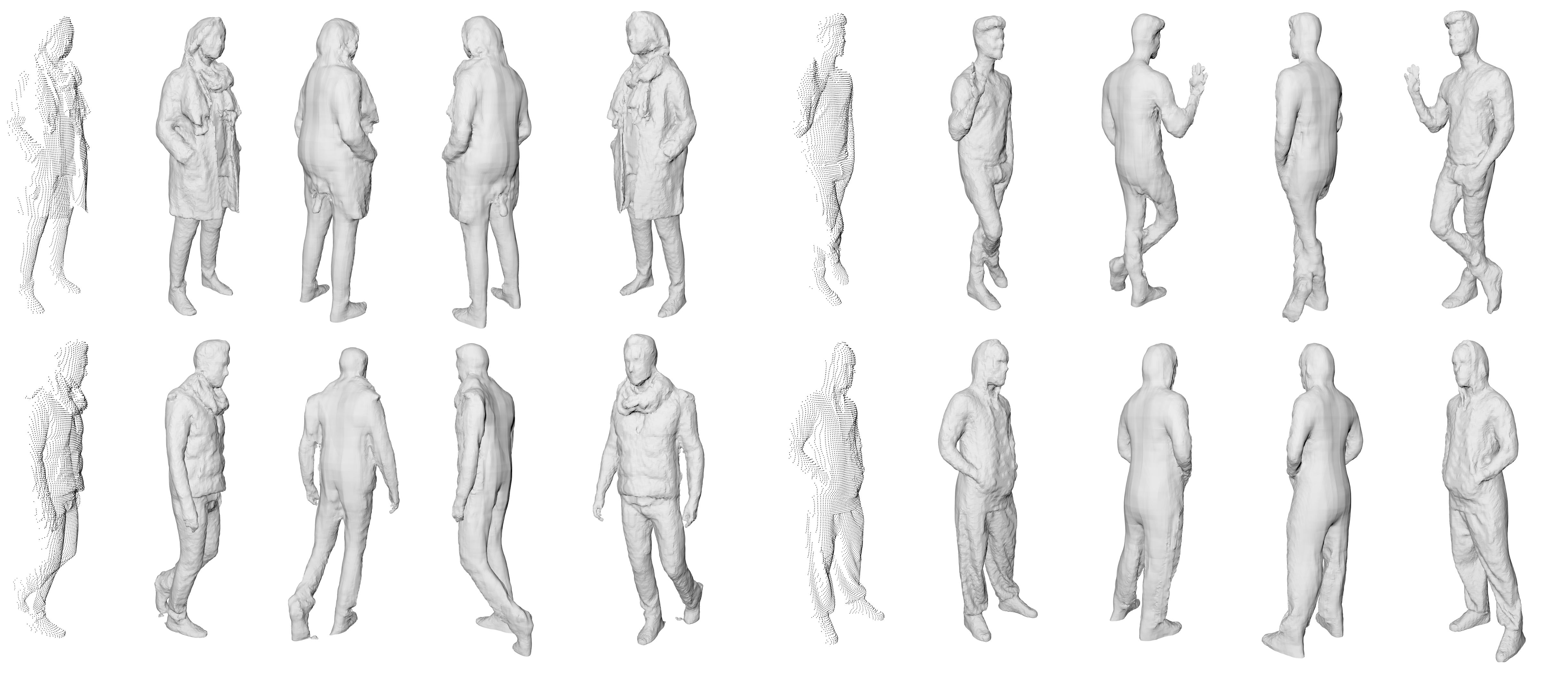}
\vspace{-0.5mm}
\caption{3D single view reconstructions from point clouds (note that the back is completely occluded). For four different single-view point clouds, we show our reconstructions from four different viewpoints.} 
\label{fig:pc_reconstruction}
\vspace{-2mm}
\end{figure*}
As a second task, we apply our method to 3D super-resolution.
To effectively solve this task, our method needs to again preserve the input shape while reconstructing details not present in the input.
Our results in side-by-side comparison with the baselines are depicted in Fig.~\ref{fig:shapenet} (bottom).
While most baseline methods either hallucinate structure or completely fail, our method consistently produces accurate and highly detailed results.
This is also reflected in the numerical comparison in Tab.~\ref{tab:shapenet_numbers_voxels}, where we improve over the baselines in all metrics.

The two last examples in Fig.~\ref{fig:shapenet} illustrate the limitations of current implicit methods: If a shape differs too much from the training set, the method fails or seems to return a similar previously seen example.
Consequentially, we hypothesize that the current methods are not suited for tasks where classification into shape prototypes is not sufficient.
This is for example the case for humans as they come in various shapes and articulations.
To verify our hypothesis, we additionally perform 3D super-resolution on our Humans dataset.
Here the advantages are even more prominent:
Our method is the only one that consistently reconstructs all limbs and produces highly detailed results.
Implicit learning-based baselines produce truncated or completely missing limbs.
We outperform all baselines also quantitatively (see Tab~\ref{tab:human_numbers_voxels}).



\subsection{Single-View Human Reconstruction}
\label{subsec:svhr}
Finally, to demonstrate the full capabilities of IF-Nets, we use them for single-view human reconstruction.
In this task, only a partial 3D point cloud is given as input -- the typical output of a depth camera.
We conduct this experiment on the challenging Humans dataset, by rendering a $250\times250$ resolution depth image, yielding around 5000 points on the visible side of the subject.
To successfully fulfill this task, our model has to simultaneously reconstruct novel articulations, retain fine details, and complete the missing data at the occluded regions -- the input contains only one side of the underlying shape.
Despite these challenges, our model is capable of reconstructing plausible and highly detailed shapes.
In Fig.~\ref{fig:pc_reconstruction}, we show both input and our results from four different angles.
Note how fine structures like the scarfs, wrinkles, or individual fingers are present in the reconstructed shapes.
Although the backside region (occluded) has less details than the visible one,
IF-Nets always produce plausible surfaces.    

This can also be seen quantitatively. \emph{Ours}: IoU 0.86, Chamfer-$L_2$  $0.011\times 10^{-2}$, Normal-Consistency 0.90. \emph{Input point cloud}:  Chamfer-$L_2$ $0.252\times 10^{-2}$. The quantitative results are in between the reconstruction quality of $32^3$ and $128^3$ full subject voxel inputs (see Tab. \ref{tab:human_numbers_voxels}), which once more validates that IF-Nets can complete single-view data.
In the supplementary video, we show an additional result on single-view reconstruction on the BUFF dataset~\cite{mh_shapeundercloth:CVPR17} from \textit{video} (without retraining nor fine tuning the model).

\section{Discussion and Conclusion}
\label{sec:conclusion}
In this work, we have introduced IF-Nets for 3D reconstruction and completion from deficient 3D inputs.
First, we have argued for an encoding consisting of a 3D multi-scale tensor of deep features, which is aligned with the Euclidean space embedding the shape. Second, instead of classifying x-y-z coordinates directly, we classify deep features extracted at their location. 
Experiments demonstrate that IF-Nets deliver continuous outputs, can reconstruct multiple topologies such as 3D humans in varied clothing, and 3D objects from ShapeNet. Quantitatively, IF-Nets outperform all state-of-the-art baselines by a large margin in all tasks.
Our reconstruction from single-view point clouds(detailed on the visible part but with missing data on the occluded part), demonstrate the strengths of 
IF-Nets: details in the input are preserved, while the shape is \emph{completed} on the occluded part, even for articulated shapes. 

Future work will explore extending IF-Nets to be generative, that is being able to sample detailed hypothesis conditioned on partial input. We also plan to address image-based reconstruction in 2 stages: first predicting a depth map, and then completing shape with IF-Nets. 

With a rising number of computer vision image reconstruction methods producing partial 3D point clouds and voxels, and 3D scanners and depth cameras becoming accessible, 3D (deficient and incomplete) data will be omnipresent in the future, and IF-Nets have the potential to be an important building block for its reconstruction and completion. 

\noindent
\begin{minipage}{\columnwidth}
    \vspace{2mm}
    \footnotesize
	\noindent
	\textbf{Acknowledgments.}~
	\hyphenation{For-schungs-ge-mein-schaft}
We would like to thank Verica Lazova for helping creating the figures, Bharat Lal Bhatnagar for helping with data preprocessing, and Lars Mescheder for sharing their watertight Shapenet meshes. This is work is funded by the Deutsche Forschungsgemeinschaft (DFG, German Research Foundation) - 409792180 (Emmy Noether Programme, project: Real Virtual Humans). We would like to thank Twindom for providing us with the scan data.
\end{minipage}

{\small
	\bibliographystyle{ieee_fullname}
	\bibliography{egbib}

\begin{thebibliography}{10}\itemsep=-1pt

\bibitem{mh_alldieck19cvpr}
Thiemo Alldieck, Marcus Magnor, Bharat~Lal Bhatnagar, Christian Theobalt, and
  Gerard Pons-Moll.
\newblock Learning to reconstruct people in clothing from a single {RGB}
  camera.
\newblock In {\em {IEEE} Conference on Computer Vision and Pattern
  Recognition}, 2019.

\bibitem{mh_alldieck2018detailed}
Thiemo Alldieck, Marcus Magnor, Weipeng Xu, Christian Theobalt, and Gerard
  Pons-Moll.
\newblock Detailed human avatars from monocular video.
\newblock In {\em International Conference on 3D Vision}, 2018.

\bibitem{mh_alldieck2018video}
Thiemo Alldieck, Marcus Magnor, Weipeng Xu, Christian Theobalt, and Gerard
  Pons-Moll.
\newblock Video based reconstruction of {3D} people models.
\newblock In {\em {IEEE} Conference on Computer Vision and Pattern
  Recognition}, 2018.

\bibitem{mh_alldieck2019tex2shape}
Thiemo Alldieck, Gerard Pons-Moll, Christian Theobalt, and Marcus Magnor.
\newblock Tex2shape: Detailed full human body geometry from a single image.
\newblock In {\em {IEEE} International Conference on Computer Vision}. {IEEE},
  2019.

\bibitem{ph_aumentado2019geometric}
Tristan Aumentado-Armstrong, Stavros Tsogkas, Allan Jepson, and Sven Dickinson.
\newblock Geometric disentanglement for generative latent shape models.
\newblock In {\em Proceedings of the IEEE International Conference on Computer
  Vision}, pages 8181--8190, 2019.

\bibitem{c_ball_pivoting}
Fausto Bernardini, Joshua Mittleman, Holly~E. Rushmeier, Cl{\'{a}}udio~T.
  Silva, and Gabriel Taubin.
\newblock The ball-pivoting algorithm for surface reconstruction.
\newblock {\em {IEEE} Transactions on Visualization and Computer Graphics},
  5(4):349--359, 1999.

\bibitem{mh_bhatnagar2019mgn}
Bharat~Lal Bhatnagar, Garvita Tiwari, Christian Theobalt, and Gerard Pons-Moll.
\newblock Multi-garment net: Learning to dress 3d people from images.
\newblock In {\em {IEEE} International Conference on Computer Vision}. {IEEE},
  2019.

\bibitem{c_Calakli11}
Fatih Calakli and Gabriel Taubin.
\newblock {SSD:} smooth signed distance surface reconstruction.
\newblock {\em Computer Graphics Forum}, 30(7):1993--2002, 2011.

\bibitem{shapenet}
Angel~X Chang, Thomas Funkhouser, Leonidas Guibas, Pat Hanrahan, Qixing Huang,
  Zimo Li, Silvio Savarese, Manolis Savva, Shuran Song, Hao Su, Jianxiong Xiao,
  Li Yi, and Fisher Yu.
\newblock Shapenet: An information-rich 3d model repository.
\newblock {\em arXiv preprint arXiv:1512.03012}, 2015.

\bibitem{i_IMGAN19}
Zhiqin Chen and Hao Zhang.
\newblock Learning implicit fields for generative shape modeling.
\newblock In {\em {IEEE} Conference on Computer Vision and Pattern
  Recognition}, pages 5939--5948, 2019.

\bibitem{v_Choy16}
Christopher~Bongsoo Choy, Danfei Xu, JunYoung Gwak, Kevin Chen, and Silvio
  Savarese.
\newblock 3d-r2n2: {A} unified approach for single and multi-view 3d object
  reconstruction.
\newblock In {\em European Conference on Computer Vision}, pages 628--644,
  2016.

\bibitem{v_TSDF_CurlessL96}
Brian Curless and Marc Levoy.
\newblock A volumetric method for building complex models from range images.
\newblock In {\em Computer Graphics and Interactive Techniques}, pages
  303--312, 1996.

\bibitem{m_Dai19}
Angela Dai and Matthias Nie{\ss}ner.
\newblock Scan2mesh: From unstructured range scans to 3d meshes.
\newblock In {\em {IEEE} Conference on Computer Vision and Pattern
  Recognition}, pages 5574--5583, 2019.

\bibitem{v_Dai17}
Angela Dai, Charles~Ruizhongtai Qi, and Matthias Nie{\ss}ner.
\newblock Shape completion using 3d-encoder-predictor cnns and shape synthesis.
\newblock In {\em {IEEE} Conference on Computer Vision and Pattern
  Recognition}, pages 6545--6554, 2017.

\bibitem{p_Fan17}
Haoqiang Fan, Hao Su, and Leonidas~J. Guibas.
\newblock A point set generation network for 3d object reconstruction from a
  single image.
\newblock In {\em {IEEE} Conference on Computer Vision and Pattern
  Recognition}, pages 2463--2471, 2017.

\bibitem{vh_Rogez}
Valentin Gabeur, Jean{-}S{\'{e}}bastien Franco, Xavier Martin, Cordelia Schmid,
  and Gr{\'{e}}gory Rogez.
\newblock Moulding humans: Non-parametric 3d human shape estimation from single
  images.
\newblock In {\em {IEEE} International Conference on Computer Vision}, pages
  2232--2241, 2019.

\bibitem{vh_Gilbert}
Andrew Gilbert, Marco Volino, John~P. Collomosse, and Adrian Hilton.
\newblock Volumetric performance capture from minimal camera viewpoints.
\newblock In {\em European Conference on Computer Vision}, pages 591--607,
  2018.

\bibitem{v_Girdhar16}
Rohit Girdhar, David~F. Fouhey, Mikel Rodriguez, and Abhinav Gupta.
\newblock Learning a predictable and generative vector representation for
  objects.
\newblock In {\em European Conference on Computer Vision}, pages 484--499,
  2016.

\bibitem{meshRCNN}
Georgia Gkioxari, Jitendra Malik, and Justin Johnson.
\newblock Mesh {R-CNN}.
\newblock In {\em {IEEE} International Conference on Computer Vision}, pages
  9785--9795, 2019.

\bibitem{m_Groueix18}
Thibault Groueix, Matthew Fisher, Vladimir~G. Kim, Bryan~C. Russell, and
  Mathieu Aubry.
\newblock A papier-m{\^{a}}ch{\'{e}} approach to learning 3d surface
  generation.
\newblock In {\em {IEEE} Conference on Computer Vision and Pattern
  Recognition}, pages 216--224, 2018.

\bibitem{mh_habermann2019TOG}
Marc Habermann, Weipeng Xu, , Michael Zollhoefer, Gerard Pons-Moll, and
  Christian Theobalt.
\newblock Livecap: Real-time human performance capture from monocular video.
\newblock {\em ACM Transactions on Graphics, (Proc. SIGGRAPH)}, jul 2019.

\bibitem{mh_habermann20deepcap}
Marc Habermann, Weipeng Xu, , Michael Zollhoefer, Gerard Pons-Moll, and
  Christian Theobalt.
\newblock Deepcap: Monocular human performance capture using weak supervision.
\newblock In {\em {IEEE} Conference on Computer Vision and Pattern Recognition
  (CVPR)}. {IEEE}, jun 2020.

\bibitem{3Dobject_survey}
Xian{-}Feng Han, Hamid Laga, and Mohammed Bennamoun.
\newblock Image-based 3d object reconstruction: State-of-the-art and trends in
  the deep learning era.
\newblock {\em {IEEE} Transactions on Pattern Analysis and Machine
  Intelligence}, 2019.

\bibitem{v_Hane17}
Christian Hane, Shubham Tulsiani, and Jitendra Malik.
\newblock Hierarchical surface prediction for 3d object reconstruction.
\newblock In {\em International Conference on 3D Vision}, pages 412--420, 2017.

\bibitem{p_InsafutdinovD18}
Eldar Insafutdinov and Alexey Dosovitskiy.
\newblock Unsupervised learning of shape and pose with differentiable point
  clouds.
\newblock In {\em Advances in Neural Information Processing Systems}, pages
  2807--2817, 2018.

\bibitem{v_Mengqi17}
Mengqi Ji, Juergen Gall, Haitian Zheng, Yebin Liu, and Lu Fang.
\newblock Surfacenet: An end-to-end 3d neural network for multiview stereopsis.
\newblock In {\em {IEEE} International Conference on Computer Vision}, pages
  2326--2334, 2017.

\bibitem{p_Jiang}
Haiyong Jiang, Jianfei Cai, and Jianmin Zheng.
\newblock Skeleton-aware 3d human shape reconstruction from point clouds.
\newblock In {\em {IEEE} International Conference on Computer Vision}, pages
  5431--5441, 2019.

\bibitem{mh_kanazawa2018endtoend}
Angjoo Kanazawa, Michael~J. Black, David~W. Jacobs, and Jitendra Malik.
\newblock End-to-end recovery of human shape and pose.
\newblock In {\em {IEEE} Conference on Computer Vision and Pattern
  Recognition}. IEEE Computer Society, 2018.

\bibitem{v_Kar17}
Abhishek Kar, Christian H{\"{a}}ne, and Jitendra Malik.
\newblock Learning a multi-view stereo machine.
\newblock In {\em Advances in Neural Information Processing Systems}, pages
  365--376, 2017.

\bibitem{c_poission}
Michael~M. Kazhdan, Matthew Bolitho, and Hugues Hoppe.
\newblock Poisson surface reconstruction.
\newblock In {\em Eurographics Symposium on Geometry Processing}, pages 61--70,
  2006.

\bibitem{c_poission2}
Michael~M. Kazhdan and Hugues Hoppe.
\newblock Screened poisson surface reconstruction.
\newblock {\em {ACM} Transactions on Graphics}, 32(3):29:1--29:13, 2013.

\bibitem{mh_Pavlakos18}
Nikos Kolotouros, Georgios Pavlakos, Michael~J. Black, and Kostas Daniilidis.
\newblock Learning to reconstruct 3d human pose and shape via model-fitting in
  the loop.
\newblock In {\em {IEEE} International Conference on Computer Vision}, pages
  2252--2261, 2019.

\bibitem{mh_Kolotouros19}
Nikos Kolotouros, Georgios Pavlakos, and Kostas Daniilidis.
\newblock Convolutional mesh regression for single-image human shape
  reconstruction.
\newblock In {\em {IEEE} Conference on Computer Vision and Pattern
  Recognition}, pages 4501--4510, 2019.

\bibitem{imagenet}
Alex Krizhevsky, Ilya Sutskever, and Geoffrey~E. Hinton.
\newblock Imagenet classification with deep convolutional neural networks.
\newblock In {\em Advances in Neural Information Processing Systems}, pages
  1106--1114, 2012.

\bibitem{i_StructuredImplicitFunctions}
Daniel Vlasic Aaron Sarna William T. Freeman Thomas~Funkhouser Kyle~Genova,
  Forrester~Cole.
\newblock Learning shape templates with structured implicit functions.
\newblock Nov 2019.

\bibitem{v_Ladicky17}
Lubor Ladicky, Olivier Saurer, SoHyeon Jeong, Fabio Maninchedda, and Marc
  Pollefeys.
\newblock From point clouds to mesh using regression.
\newblock In {\em {IEEE} International Conference on Computer Vision}, pages
  3913--3922, 2017.

\bibitem{mh_lazova3dv2019}
Verica Lazova, Eldar Insafutdinov, and Gerard Pons-Moll.
\newblock 360-degree textures of people in clothing from a single image.
\newblock In {\em International Conference on 3D Vision (3DV)}, sep 2019.

\bibitem{vh_Leroy18}
Vincent Leroy, Jean{-}S{\'{e}}bastien Franco, and Edmond Boyer.
\newblock Shape reconstruction using volume sweeping and learned
  photoconsistency.
\newblock In {\em European Conference on Computer Vision}, pages 796--811,
  2018.

\bibitem{v_Liao}
Yiyi Liao, Simon Donn{\'{e}}, and Andreas Geiger.
\newblock Deep marching cubes: Learning explicit surface representations.
\newblock In {\em {IEEE} Conference on Computer Vision and Pattern
  Recognition}, pages 2916--2925, 2018.

\bibitem{v_Liu18}
Shikun Liu, Lee Giles, and Alexander Ororbia.
\newblock Learning a hierarchical latent-variable model of 3d shapes.
\newblock In {\em International Conference on 3D Vision}, pages 542--551. IEEE,
  2018.

\bibitem{mh_smpl2015loper}
Matthew Loper, Naureen Mahmood, Javier Romero, Gerard Pons-Moll, and Michael~J
  Black.
\newblock {SMPL}: A skinned multi-person linear model.
\newblock {\em {ACM} Transactions on Graphics}, 34(6):248:1--248:16, 2015.

\bibitem{c_marching_cubes}
William~E. Lorensen and Harvey~E. Cline.
\newblock Marching cubes: {A} high resolution 3d surface construction
  algorithm.
\newblock In {\em Computer Graphics and Interactive Techniques}, pages
  163--169, 1987.

\bibitem{i_OccNet19}
Lars~M. Mescheder, Michael Oechsle, Michael Niemeyer, Sebastian Nowozin, and
  Andreas Geiger.
\newblock Occupancy networks: Learning 3d reconstruction in function space.
\newblock In {\em {IEEE} Conference on Computer Vision and Pattern
  Recognition}, pages 4460--4470, 2019.

\bibitem{i_DeepLevelSet}
Mateusz Michalkiewicz, Jhony~K Pontes, Dominic Jack, Mahsa Baktashmotlagh, and
  Anders Eriksson.
\newblock Deep level sets: Implicit surface representations for 3d shape
  inference.
\newblock {\em arXiv preprint arXiv:1901.06802}, 2019.

\bibitem{ih_Dynamic_Fusion}
Richard~A. Newcombe, Dieter Fox, and Steven~M. Seitz.
\newblock Dynamicfusion: Reconstruction and tracking of non-rigid scenes in
  real-time.
\newblock In {\em {IEEE} Conference on Computer Vision and Pattern
  Recognition}, pages 343--352, 2015.

\bibitem{v_olszewski2019tbn}
Kyle Olszewski, Sergey Tulyakov, Oliver Woodford, Hao Li, and Linjie Luo.
\newblock Transformable bottleneck networks.
\newblock {\em The IEEE International Conference on Computer Vision (ICCV)},
  Nov 2019.

\bibitem{mh_omran2018neural}
Mohamed Omran, Christop Lassner, Gerard Pons-Moll, Peter Gehler, and Bernt
  Schiele.
\newblock Neural body fitting: Unifying deep learning and model based human
  pose and shape estimation.
\newblock In {\em International Conference on 3D Vision}, 2018.

\bibitem{i_DeepSDF}
Jeong~Joon Park, Peter Florence, Julian Straub, Richard~A. Newcombe, and Steven
  Lovegrove.
\newblock Deepsdf: Learning continuous signed distance functions for shape
  representation.
\newblock In {\em {IEEE} Conference on Computer Vision and Pattern
  Recognition}, pages 165--174, 2019.

\bibitem{mh_patel2020}
Chaitanya Patel, Zhouyingcheng Liao, and Gerard Pons-Moll.
\newblock Tailornet: Predicting clothing in 3d as a function of human pose,
  shape and garment style.
\newblock In {\em {IEEE} Conference on Computer Vision and Pattern Recognition
  (CVPR)}. {IEEE}, jun 2020.

\bibitem{p_Pauly}
Mark Pauly, Markus~H. Gross, and Leif Kobbelt.
\newblock Efficient simplification of point-sampled surfaces.
\newblock In {\em {IEEE} Visualization}, pages 163--170, 2002.

\bibitem{mh_ponsmoll2017clothcap}
Gerard Pons-Moll, Sergi Pujades, Sonny Hu, and Michael Black.
\newblock {ClothCap}: Seamless {4D} clothing capture and retargeting.
\newblock {\em {ACM} Transactions on Graphics}, 36(4), 2017.

\bibitem{p_humans_Basis}
Sergey Prokudin, Christoph Lassner, and Javier Romero.
\newblock Efficient learning on point clouds with basis point sets.
\newblock In {\em {IEEE} International Conference on Computer Vision
  Workshops}, 2019.

\bibitem{mh_Pumarola}
Albert Pumarola, Jordi Sanchez, Gary P.~T. Choi, Alberto Sanfeliu, and Francesc
  Moreno{-}Noguer.
\newblock 3dpeople: Modeling the geometry of dressed humans.
\newblock In {\em {IEEE} International Conference on Computer Vision}, pages
  2242--2251, 2019.

\bibitem{p_PointNet}
Charles~Ruizhongtai Qi, Hao Su, Kaichun Mo, and Leonidas~J. Guibas.
\newblock Pointnet: Deep learning on point sets for 3d classification and
  segmentation.
\newblock In {\em {IEEE} Conference on Computer Vision and Pattern
  Recognition}, pages 77--85, 2017.

\bibitem{p_PointNet++}
Charles~Ruizhongtai Qi, Hao Su, Kaichun Mo, and Leonidas~J. Guibas.
\newblock Pointnet: Deep learning on point sets for 3d classification and
  segmentation.
\newblock In {\em {IEEE} Conference on Computer Vision and Pattern
  Recognition}, pages 77--85, 2017.

\bibitem{m_Ranjan18}
Anurag Ranjan, Timo Bolkart, Soubhik Sanyal, and Michael~J. Black.
\newblock Generating 3d faces using convolutional mesh autoencoders.
\newblock In {\em European Conference on Computer Vision}, pages 725--741,
  2018.

\bibitem{v_Jimenez16}
Danilo~Jimenez Rezende, S.~M.~Ali Eslami, Shakir Mohamed, Peter~W. Battaglia,
  Max Jaderberg, and Nicolas Heess.
\newblock Unsupervised learning of 3d structure from images.
\newblock In {\em Advances in Neural Information Processing Systems}, pages
  4997--5005, 2016.

\bibitem{v_Riegler17}
Gernot Riegler, Ali~Osman Ulusoy, Horst Bischof, and Andreas Geiger.
\newblock Octnetfusion: Learning depth fusion from data.
\newblock In {\em International Conference on 3D Vision}, pages 57--66, 2017.

\bibitem{ih_PiFu}
Shunsuke Saito, Zeng Huang, Ryota Natsume, Shigeo Morishima, Angjoo Kanazawa,
  and Hao Li.
\newblock Pifu: Pixel-aligned implicit function for high-resolution clothed
  human digitization.
\newblock In {\em {IEEE} International Conference on Computer Vision}, pages
  2304--2314, 2019.

\bibitem{p_treeGAN}
Dong~Wook Shu, Sung~Woo Park, and Junseok Kwon.
\newblock 3d point cloud generative adversarial network based on tree
  structured graph convolutions.
\newblock In {\em {IEEE} International Conference on Computer Vision}, pages
  3859--3868, 2019.

\bibitem{ih_KillingFusion}
Miroslava Slavcheva, Maximilian Baust, Daniel Cremers, and Slobodan Ilic.
\newblock Killingfusion: Non-rigid 3d reconstruction without correspondences.
\newblock In {\em {IEEE} Conference on Computer Vision and Pattern
  Recognition}, 2017.

\bibitem{vh_facsimile}
David Smith, Matthew Loper, Xiaochen Hu, Paris Mavroidis, and Javier Romero.
\newblock {FACSIMILE:} fast and accurate scans from an image in less than a
  second.
\newblock In {\em {IEEE} International Conference on Computer Vision}, 2019.

\bibitem{v_Smith18}
Edward Smith, Scott Fujimoto, and David Meger.
\newblock Multi-view silhouette and depth decomposition for high resolution 3d
  object representation.
\newblock In S. Bengio, H. Wallach, H. Larochelle, K. Grauman, N. Cesa-Bianchi,
  and R. Garnett, editors, {\em Advances in Neural Information Processing
  Systems 31}, pages 6479--6489. Curran Associates, Inc., 2018.

\bibitem{v_stutz}
David Stutz and Andreas Geiger.
\newblock Learning 3d shape completion from laser scan data with weak
  supervision.
\newblock In {\em {IEEE} Conference on Computer Vision and Pattern
  Recognition}, pages 1955--1964, 2018.

\bibitem{v_Tatarchenko17}
Maxim Tatarchenko, Alexey Dosovitskiy, and Thomas Brox.
\newblock Octree generating networks: Efficient convolutional architectures for
  high-resolution 3d outputs.
\newblock In {\em {IEEE} International Conference on Computer Vision}, pages
  2107--2115, 2017.

\bibitem{v_ShapeReconstructionIsClassification}
Maxim Tatarchenko, Stephan~R. Richter, Ren{\'{e}} Ranftl, Zhuwen Li, Vladlen
  Koltun, and Thomas Brox.
\newblock What do single-view 3d reconstruction networks learn?
\newblock In {\em {IEEE} Conference on Computer Vision and Pattern
  Recognition}, pages 3405--3414, 2019.

\bibitem{p_Kernel}
Hugues Thomas, Charles~R. Qi, Jean{-}Emmanuel Deschaud, Beatriz Marcotegui,
  Fran{\c{c}}ois Goulette, and Leonidas~J. Guibas.
\newblock Kpconv: Flexible and deformable convolution for point clouds.
\newblock In {\em {IEEE} International Conference on Computer Vision}, pages
  4541--4550, 2019.

\bibitem{v_Tulsiani17}
Shubham Tulsiani, Tinghui Zhou, Alexei~A. Efros, and Jitendra Malik.
\newblock Multi-view supervision for single-view reconstruction via
  differentiable ray consistency.
\newblock In {\em {IEEE} Conference on Computer Vision and Pattern
  Recognition}, pages 209--217, 2017.

\bibitem{mh_tung2017self}
Hsiao-Yu Tung, Hsiao-Wei Tung, Ersin Yumer, and Katerina Fragkiadaki.
\newblock Self-supervised learning of motion capture.
\newblock In {\em Advances in Neural Information Processing Systems}, pages
  5236--5246, 2017.

\bibitem{vh_varol2018bodynet}
G{\"u}l Varol, Duygu Ceylan, Bryan Russell, Jimei Yang, Ersin Yumer, Ivan
  Laptev, and Cordelia Schmid.
\newblock Bodynet: Volumetric inference of 3d human body shapes.
\newblock In {\em European Conference on Computer Vision}, 2018.

\bibitem{v_local_global}
Hao Wang, Nadav Schor, Ruizhen Hu, Haibin Huang, Daniel Cohen{-}Or, and Hui
  Huang.
\newblock Global-to-local generative model for 3d shapes.
\newblock {\em {ACM} Transactions on Graphics}, 37(6):214:1--214:10, 2018.

\bibitem{m_Wang18}
Nanyang Wang, Yinda Zhang, Zhuwen Li, Yanwei Fu, Wei Liu, and Yu{-}Gang Jiang.
\newblock Pixel2mesh: Generating 3d mesh models from single {RGB} images.
\newblock In {\em European Conference on Computer Vision}, pages 55--71, 2018.

\bibitem{v_Wu17}
Jiajun Wu, Yifan Wang, Tianfan Xue, Xingyuan Sun, Bill Freeman, and Josh
  Tenenbaum.
\newblock Marrnet: 3d shape reconstruction via 2.5d sketches.
\newblock In {\em Advances in Neural Information Processing Systems}, pages
  540--550, 2017.

\bibitem{v_Wu16}
Jiajun Wu, Chengkai Zhang, Tianfan Xue, Bill Freeman, and Josh Tenenbaum.
\newblock Learning a probabilistic latent space of object shapes via 3d
  generative-adversarial modeling.
\newblock In {\em Advances in Neural Information Processing Systems}, pages
  82--90, 2016.

\bibitem{v_Wu15}
Zhirong Wu, Shuran Song, Aditya Khosla, Fisher Yu, Linguang Zhang, Xiaoou Tang,
  and Jianxiong Xiao.
\newblock 3d shapenets: {A} deep representation for volumetric shapes.
\newblock In {\em {IEEE} Conference on Computer Vision and Pattern
  Recognition}, pages 1912--1920, 2015.

\bibitem{i_DISN}
Qiangeng Xu, Weiyue Wang, Duygu Ceylan, Radom{\'{\i}}r Mech, and Ulrich
  Neumann.
\newblock {DISN:} deep implicit surface network for high-quality single-view 3d
  reconstruction.
\newblock In {\em Advances in Neural Information Processing Systems}, pages
  490--500, 2019.

\bibitem{p_PointFlow}
Guandao Yang, Xun Huang, Zekun Hao, Ming{-}Yu Liu, Serge~J. Belongie, and
  Bharath Hariharan.
\newblock Pointflow: 3d point cloud generation with continuous normalizing
  flows.
\newblock In {\em {IEEE} International Conference on Computer Vision}, pages
  4541--4550, 2019.

\bibitem{ih_DoubleFusion}
Tao Yu, Zerong Zheng, Kaiwen Guo, Jianhui Zhao, Qionghai Dai, Hao Li, Gerard
  Pons{-}Moll, and Yebin Liu.
\newblock Doublefusion: Real-time capture of human performances with inner body
  shapes from a single depth sensor.
\newblock In {\em {IEEE} Conference on Computer Vision and Pattern
  Recognition}, pages 7287--7296, 2018.

\bibitem{mh_zanfir2018monocular}
Andrei Zanfir, Elisabeta Marinoiu, and Cristian Sminchisescu.
\newblock Monocular 3d pose and shape estimation of multiple people in natural
  scenes--the importance of multiple scene constraints.
\newblock In {\em {IEEE} Conference on Computer Vision and Pattern
  Recognition}, pages 2148--2157, 2018.

\bibitem{mh_shapeundercloth:CVPR17}
Chao Zhang, Sergi Pujades, Michael Black, and Gerard Pons-Moll.
\newblock Detailed, accurate, human shape estimation from clothed {3D} scan
  sequences.
\newblock In {\em {IEEE} Conf. on Computer Vision and Pattern Recognition
  (CVPR)}, 2017.

\bibitem{v_Zhang18}
Xiuming Zhang, Zhoutong Zhang, Chengkai Zhang, Josh Tenenbaum, Bill Freeman,
  and Jiajun Wu.
\newblock Learning to reconstruct shapes from unseen classes.
\newblock In {\em Advances in Neural Information Processing Systems}, pages
  2263--2274, 2018.

\bibitem{vh_DeepHumans}
Zerong Zheng, Tao Yu, Yixuan Wei, Qionghai Dai, and Yebin Liu.
\newblock Deephuman: 3d human reconstruction from a single image.
\newblock In {\em {IEEE} International Conference on Computer Vision}, pages
  7739--7749, 2019.

\end{thebibliography}
}
\end{document}